\documentclass[a4paper,11pt,oneside]{report}
\usepackage[left=2cm,right=2cm,top=2cm,bottom=3cm]{geometry}

\usepackage{amsmath}
\usepackage{graphicx}
\usepackage{verbatim}
\usepackage{latexsym}
\usepackage{mathchars}
\usepackage{setspace}

\usepackage{algorithm}
\usepackage[noend]{algpseudocode}
\makeatletter
\def\BState{\State\hskip-\ALG@thistlm}
\makeatother

\input{blocked.sty}
\input{uhead.sty}
\input{boxit.sty}
\input{icthesis.sty}

\newcommand{\norm}[1]{\left\| #1 \right\|}

\newcommand{\NN}{{\sf I\kern-0.14emN}}   
\newcommand{\ZZ}{{\sf Z\kern-0.45emZ}}   
\newcommand{\QQQ}{{\sf C\kern-0.48emQ}}   
\newcommand{\RR}{{\sf I\kern-0.14emR}}   

\newcommand{\normallinespacing}{\renewcommand{\baselinestretch}{1.5} \normalsize}

\newcommand{\syncc}{~\stackrel{\textstyle \rhd\kern-0.57em\lhd}{\scriptstyle L}~}

\begin{document}

\title{\LARGE {\bf Per-Pixel Feedback for improving Semantic Segmentation}\\
 \vspace*{6mm}
}

\author{Aditya Ganeshan}
\submitdate{May 2017}

\normallinespacing
\maketitle

\preface
\cleardoublepage

\begin{declaration}
  I hereby certify that the work which is being presented in the dissertation entitled \textbf{Per-pixel Feed-back For Improving Semantic Segmentation}, in the partial fulfillment of the requirement for the award of the Degree of Integrated Master of Science in Applied Mathematics and submitted in the Department of Mathematics, Indian Institute of Technology Roorkee, is an authentic record of my own work carried out during a period from January, 2017 to April, 2017 under the supervision of \textbf{ N. Sukavanam}, Professor, Department of Mathematics, Indian Institute of Technology Roorkee and \textbf{R. Venkatesh Babu}, Associate Professor, Indian Institute of Science, Bangalore. The matter presented in this report has not been submitted by me for the award of any other degree of this or any other institute.\\
  
  Date :  \hspace{10.5cm} (Aditya Ganeshan) \\

    This is certified that the above statement made by the candidate is correct to the best of my knowledge.\\
    
    Date :  \hspace{11cm} (R. Venkatesh Babu) \\
  	\hspace*{12.5cm} Associate Professor\\ \hspace*{7.5cm} \raggedright Department of Computational and Data Sciences\\ \hspace*{13.3cm} IISc Bangalore
      
  This is certified that the above statement made by the candidate is correct to the best of my knowledge.
  
  Date :  \hspace{12cm} (N. Sukavanam) \\
  	\hspace*{14.3cm} Professor\\ \hspace*{11cm} \raggedright Department of Mathematics\\ \hspace*{13.8cm} IIT Roorkee \vspace{1cm}

\end{declaration}

\clearpage

\begin{abstract}

Semantic segmentation is the task of assigning a label to each pixel in the image.In recent years, deep convolutional neural networks have been driving advances in multiple tasks related to cognition. Although, DCNNs have resulted in unprecedented visual recognition performances,  they offer little transparency. To understand how DCNN based models work at the task of semantic segmentation, we try to analyze the DCNN models in semantic segmentation. We try to find the importance of global image information for labeling pixels. 

Based on the experiments on discriminative regions, and modeling of fixations, we propose a set of new training loss functions for fine-tuning DCNN based models. The proposed training regime has shown improvement in performance of DeepLab Large FOV(VGG-16) Segmentation model for PASCAL VOC 2012 dataset. However, further test remains to conclusively evaluate the benefits due to the proposed loss functions across models, and data-sets.

\end{abstract}

\addcontentsline{toc}{chapter}{Acknowledgements}

\begin{acknowledgements}

I express my deepest gratitude to my supervisors Prof. R. Venkatesh Babu, Department of Computational and Data Sciences, IISc Bangalore and Prof. N. Sukavanam, Department of Mathematics, IIT Roorkee for giving me the opportunity of working on this dissertation under their able guidance. I am extremely grateful to them for their expert opinions, valuable advice, support and continuous encouragement throughout the period in which this work was carried out.\\ \\
I would also thank my mentors at VAL, IISc Mr. Konda Reddy Mopuri and Mr. Ravi Kiran Sarvadevabhatla, under whom I was carrying out this project, for their guidance and supervision. \\ \\
Finally, I am also highly obliged to Prof. V.K. Katiyar, Head of Department of Mathematics, I.I.T. Roorkee, for providing encouragement and necessary facilities for carrying out this work. \\ \\

Place : Roorkee \hspace{10.5cm} (Aditya Ganeshan) \\
Date :

\end{acknowledgements}

\body
\chapter{Introduction}
\section*{}
One of the primary problems in computer vision is Image understanding~\cite{Hierar2013}. Image understanding has various applications such as navigation in robotics~\cite{navig}, caption generation for images~\cite{showandtell},automatic driving~\cite{autodrive}, crowd counting~\cite{crowdcounting} and many more. An important part of image understanding is full-scene labeling, which is also known as semantic segmentation.

Semantic segmentation is the task of assigning a label to each pixel in the image.It has been cited as one of the most important and challenging problem in computer vision~\cite{challenging}, as it involves detection, multi-label recognition and segmentation at the same time.Semantic segmentation can help in getting finer details of detected objects in an image, such as location, shape and size. 

The task of assigning a label to a pixel in the image relies on not only the information that is local, such as color and position but also global context such as presence of more discriminative evidence for presence of object.The task grows increasingly challenging with increase in the number of labels. Another difficult arises by properties of objects in images such as occlusion, deformation, background clutter, intra-class variation, viewpoint variation etc. A successful model must be robust to such properties. For properly segmenting the given image it is crucial to ensure the self consistency of the interpretation by the contextual information that the model/algorithm utilizes.~\cite{Hierar2013} 
  
  	\begin{figure}[thb]
		\label{fig:Semantic Segmentation}
		\centering
		\includegraphics[scale=0.45]{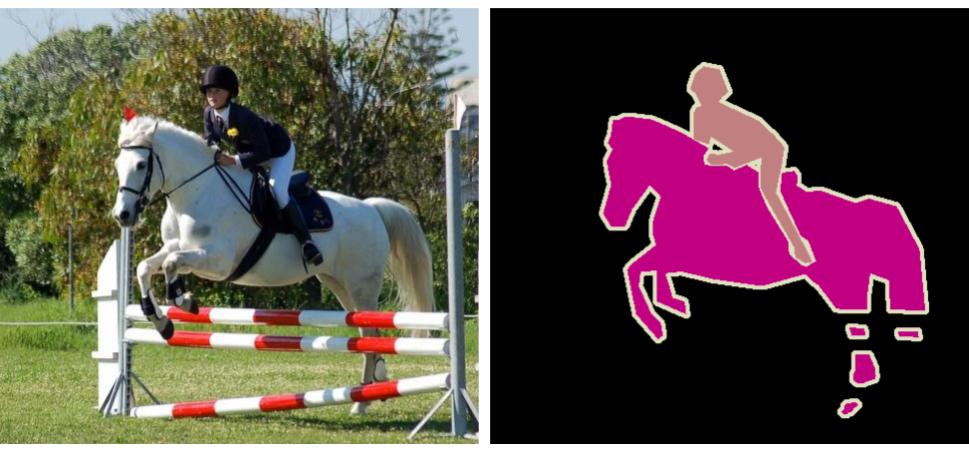}
        \caption[Semantic Segmentation Example]{An Example of Semantic segmentation}
	\end{figure}
In recent years, the rapid development of deep convolutional neural networks~\cite{cnn_example1,cnn_example2,cnn_example3,cnn_example4} have been driving advances in multiple tasks related to recognition.Availability of large scale datasets~\cite{dataset_imagenet} with millions of labeled examples, powerful and flexible implementation capabilities~\cite{caffe,torch,theano,lasagne}, and affordable GPU computation have enabled the current advances in visual recognition related tasks.  Various models based on  Deep convolutional neural networks, also know as ‘Convnets’ or DCNNs, provide state-of-the-art performance in a plethora of tasks such as Image classification, action classification, pose estimation, and many more. Similarly in the task of semantic segmentation, DCNNs based models have given the state-of-the-art performance for the past 3 years~\cite{deeplab2014,adelaide2015,pspnet2016}.

Although, DCNNs have resulted in unprecedented visual recognition performances,  they offer little transparency. DCNNs shed a little light on why and how they achieve the higher performance. Due to this, they are treated as black-box model by various researchers. Another obstacle faced in the understanding of DCNNs based models is that they couple feature-extraction and classification based on extracted features into the same process, a pattern that is very different from learning models of pre-deep learning era.

To understand how DCNN based models work at the task of semantic segmentation, we locate the discriminative image regions responsible for prediction of objects in an image. We try to additionally show the importance of global discriminative features of objects of labeling of pixels which might even be far away from the location of the discriminative feature.

We propose a set of new training loss functions for fine-tuning DCNN based models. The proposed training regime has shown improvement in performance of DeepLab Large FOV(VGG-16)~\cite{deeplab2014}. However, further test remains to conclusively evaluate the benefits due to the proposed loss functions across models, and data-sets. 

Future work will be focused on evaluating the performance of DCNN based models fine-tuned using these losses across multiple data-sets.  

\section{Preliminaries}
\subsection{Deep Convolutional Neural Networks}
\subsubsection{History}
In the human ambition to build computer systems which could stimulate the brain lies the origin of deep learning and artificial intelligence. A recent paper~\cite{dlorigins} links the modern day deep learning to the concept of ``Associationism'' by Aristotle around 300 BC. One of the earliest implementation of models close to the modern Deep learning models was The Perceptron by Frank Rosenblatt~\cite{perceptronorigin}. The perceptron is a single-layer neural network, the operation of which is based on error-correlation learning.In 1980s, Kunihiko Fukushima introduced the Neocognitron, which was a neural network model for visual pattern recognition. The Neocognitron planted seeds for deep learning models in visual pattern recognition which are now known as Convolutional Neural Networks.Various seminal works such as Hopfield Networks~\cite{hopfield1982}, Deep Boltzmann Machines~\cite{boltzmann}, LeNet~\cite{lenet} and many others have contributed to the current success that deep learning models enjoy.
\subsubsection{Why DCNNs work?}
A significant reason for the success of deep learning models was its ability to process raw natural data, where conventional machine learning algorithms failed. A computational model which is composed of multiple processing layers which learn representations of data at multiple levels of abstraction is crucial for processing raw natural data. Using the backpropagation algorithm to indicate how a machine should change its internal parameters that are used to compute the representation in each layer from the representation in the previous layer, deep learning models discovers intricate structure in large data sets~\cite{deeplecunbengiohinton}.  
\subsubsection{Deep Feed-forward Networks}
One of the fundamental deep learning models are Deep feedforward networks, also often called feedforward neural networks,or multilayer perceptrons(MLPs).A feedforward network tries to approximate some function $ f^* $. The models are called feed forward as there are no feedback connections in the model, that is, the output of the model is not feed into itself.

Neural networks are modeled as non-linear composition of linear functions. First lets look at the class of linear functions that the model uses. With an input $X$, where $X$ is a $n$ dimensional vector, and we want as an output $y$, which is an $m$ dimensional vector, then a linear function $f$ is represented as :
\begin{equation}
f(X) = W^TX +B
\end{equation}
, where $W$ is a $m \times n$ weight matrix, and $B$ is a $m$ dimensional bias vector. Thus the $i$th component of the output $y$ is a weighted linear combination of all components of the input $X$, plus a bias term $b_i$, which is the $i$th term of bias vector $B$ 

Now, The Feedforward neural network is typically represented as a composition of many such functions.Each function is represented by a layer in the model. The output of each layer acts as the input to the layer/function above. A $n$ layer network, $f^*$ having input $X^0$ and output $y$ can be represented as :
\begin{equation}
\begin{aligned}
f_1(X^0 ) &= \sigma(W^T_1 X + B_1) = X^1 \\
f_2(X^1) &= \sigma(W^T_2 X^1 + B_2) = X^2 \\
... &= ...\\
f_n(X^{n-1}) &=\sigma(W^T_{n-1} X^{n-1} + B_{n-1}) = y 
\end{aligned}
\end{equation}
where, $\sigma$ is a non-linearity function such as a $sigmoid$ or a $\tanh$,while $W^i$ and $B^i$ are the weight matrix and bias vector corresponding to each layer.
 
 	\begin{figure}[tbh]
		\label{fig:Feedforward Network}
		\centering
		\includegraphics[scale=0.40]{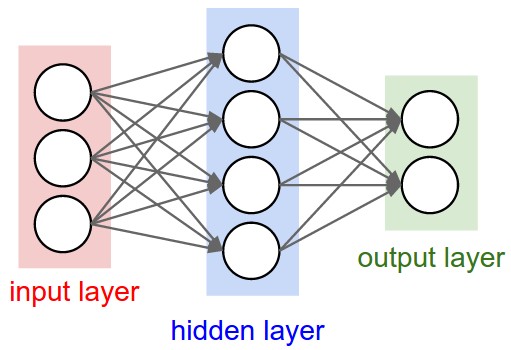}
        \caption[Feed-Forward Networks]{This figure shows the general structure of Feed-Forward Networks.}
	\end{figure}
    
One of the ways to measure the performance of our model is to consider a loss function. Consider the function we are trying to approximate to be $f^*$. We are given input $X$. Using our model we predict an output $y$. The true output can be represented as $y^* = f^*(X) $. We define a loss function or cost function $L(y,y^*)$ which acts as a measure of the error between the predicted outcome$y$ and the actual outcome $y^*$.

The total loss over a sample of data $\{ X_1,X_2,... X_n \}$ can be evaluated as, 
\begin{equation}
L_{tot} = \sum_{i = 1}^{n}{L(y_i,y_i^*)}
\end{equation}
, where $y_i$ and $y_i^*$ are the prediction and actual outcome for input $X_i$.

In deep learning, our strategy is to learn the values of all the weight matrices $W^i$ and bias vectors $B_i$ so as to minimize the total loss.Now, to learn the parameters $W_i$s and $B_i$s, we use various variations of optimization routines such as Adam~\cite{adam} , limited memory BGFS, and conjugate gradients which are based on the gradient of total loss with respect to the weights $W_i$s and Biases $B_i$s. A simple implementation of Stochastic Gradient Descent, which is a commonly used optimization routine, as given in~\cite{deeplearningbook} is given below:

\begin{algorithm}
\caption{Stochastic Gradient Descent(SGD) update at training iteration k}\label{SGD}
\begin{algorithmic}[1]
\State \textbf{Require :} learning rate $\epsilon_k$
\State \textbf{Require :} Initial parameter $\theta$ ( the parameter $W_i$ and $B_i$)
  \While{Stopping criteria is not met}
    \State Sample a minibatch of $m$ samples from the training set $\{ X_1,...,X_m \} $ with targets $y_i$
    \State Compute gradient estimate $\hat{g} \leftarrow + \frac{1}{m} \nabla_\theta \sum_i {L(y_i,y^*_i)}$
    \State Apply update $\theta \leftarrow \theta - \epsilon \hat{g}$
  \EndWhile{end While}
\end{algorithmic}
\end{algorithm}

\subsubsection{Universal approximation property}
One of the important property of neural networks is the universal approximation property~\cite{uat1,uat2}. It roughly states that an multi-layer neural network can represent any function:
\begin{itemize}
\item Boolean Approximation: an MLP of one hidden layer1 can represent any boolean
function exactly.
\item Continuous Approximation: an MLP of one hidden layer can approximate any bounded
continuous function with arbitrary accuracy.
\item Arbitrary Approximation: an MLP of two hidden layers can approximate any function
with arbitrary accuracy.
\end{itemize}

\subsubsection{Convolutional neural networks}
Convolutional neural networks~\cite{lenet,alexnet} are Variants of Feedforward neural networks for processing data that has know, grid-like topology. It has been extensively used in fields like image processing and time-series analysis. Convolutional neural networks employ a mathematical operation called `convolution' in at least one of its layers.

Convolution is a mathematical operation, on two functions which produces a third function.It is represented by the symbol $*$, and is a particular kind of Integral transform :
\begin{equation}
(f * g) (t)= \int_{-\infty}^{\infty}{f(\tau)g(t-\tau)d\tau} 
\end{equation}

In discrete settings,it can be written as 
\begin{equation}
(f * g) (t)= \sum_{\tau = -\infty}^{\infty}{f(\tau)g(t-\tau)} 
\end{equation}

For an image, which can be represented as a 2D function $f(x,y)$ we use a two dimensional convolution.With the assumption that the function $g$ that we convolve the input $f$ with is $0$ everywhere except the some finite points, we can write this operation as :
\begin{equation}
(f * g) (i,j)= \sum_{m} {\sum_{n}{f(i-m,j-n)g(m,n)} }
\end{equation}
In a convolution layer in a neural network, we perform multiple convolution operations on all possible spatial locations of the input with a fixed but learnable convolution function. The convolution function $g(x,y)$ is defined such that it is zero at all points $(x,y) \in \{(x,y)| x>m \quad or\quad y>m \}$  This function can be represented by a kernel $K_{m \times m }$ which contain values $k_{(i,j)}$ such that : 
\begin{equation}
k_{(i,j)} = g(i,j) \quad   \forall \quad 0\le i \le m,0\le j \le m 
\end{equation}
The kernel so formed is also known as a filter.Using the filter and performing convolution at all possible spatial location of the input, we get a output map $u(x,y)$ which now acts as input to the layer above it.

The key benefits of using convolution are parameter sharing, sparse interaction and equivariant representation.Due to these properties, Convolutional neural networks have been able to outperform various Feedforward neural networks at recognition task.

Convolutional neural networks are built using various kinds of layers. The most essential layers which are frequently used are : 
\begin{itemize}
\item \textbf{Convoutional Layer} :  A layer containing multiple filters, which convolve with all the spatial locations of the input. Each filter gives as an output a spatial map of values, which act as a channel in the input to the next layer.
\item \textbf{Non-linearity Layer} :  A Nonlinearity layer performs an element wise non-linearity operation on each unit of the input. Some examples of non-lienarity are ReLU~\cite{relu} and sigmoid.
\item \textbf{Pool Layer} : A pool layer performs a downsampling operation along spatial axis to reduce size of input to next layer. Additionally, downsampling by using layers such as MaxPool Layers, help to make the system invariant to small translational changes in the input~\cite{deeplearningbook}.
\item \textbf{FC Layer} : FC stands for Fully connected. In these layers, each unit is connected to all the input units, unlike connections in convolutional layers, which is sparse. Generally, these layers are stacked at the very end of the network just before the output~\cite{lenet,alexnet}.
\end{itemize}
 
 	\begin{figure}[tbh]
		\label{fig:Convolutional Network}
		\centering
		\includegraphics[scale=0.65]{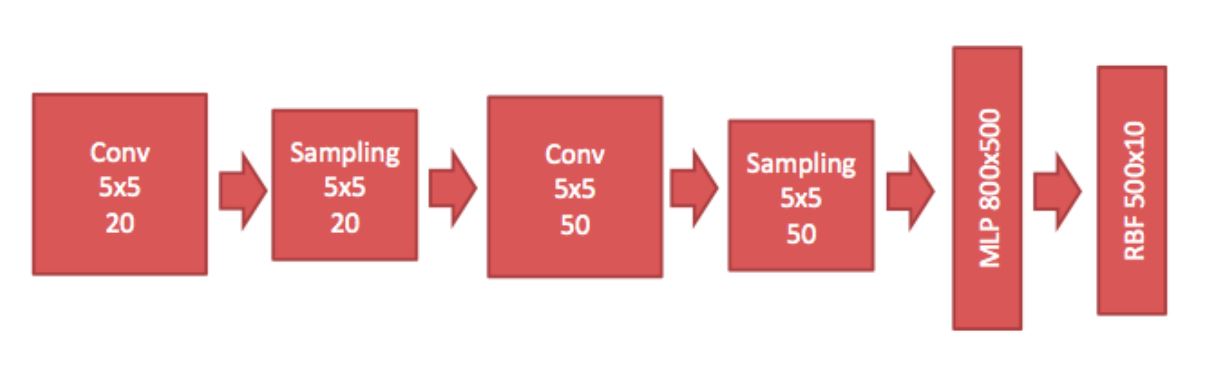}
        \caption[Convolutional Neural Network]{This figure shows the general structure of a Convolutional Neural Network called the LeNet.}
	\end{figure}

These layers are stacked over one-another to create `deep' architectures of learning models which remain end-to-end differentiable. Due to this, we are able to use optimization routines just as in the case Feedforward networks.The optimization routines depend on the loss function used for training the network. We will briefly review one of the most popular loss function used in Deep learning.

\subsubsection{Softmax-classifier Loss}

A popular choice for loss functions is the Softmax classifier.The Softmax classifier is a generalization of the binary Logistic Regression classifier to multiple classes.In the Softmax classifier,we interpret the output of the network as unnormalized log probabilities for each class.We define the cross-entropy loss as :
\begin{equation}
L_i = -log\bigg( \frac{e^{f_{y_i}}}{\sum_j e^{f_{y_j}}} \bigg)
\end{equation}
where $f_j$ stands for the $j$th element of the output $f$. The total loss $L_{tot}$, know as the data loss, is evaluated as :
\begin{equation}
L_{tot} = \frac{1}{N} \sum_{i} L_i
\end{equation}

This loss also has a probabilistic interpretation. The value $P(y_i|x_i; W)$, given by
\begin{equation}
P(y_i|x_i; W) = \frac{e^{f_{y_i}}}{\sum_j e^{f_{y_j}}}
\end{equation}
can be interpreted as the (normalized) probability assigned to the correct label $y_i$ given the image $x_i$ and parameterized by $W$. This leads us to understand why it is also called cross-entropy loss. The cross-entropy between a “true” distribution $q$ and an estimated distribution $p$ is defined as:
\begin{equation}
H(q,p) = - \sum_x{q(x) \log{p(x)}}
\end{equation}
The Softmax classifier is hence minimizing the cross-entropy between the estimated class probabilities ( $p=\frac{e^{f_{y_i}}}{\sum_j e^{f_{y_j}}}$ as seen above) and the “true” distribution, which in this interpretation is the distribution where all probability mass is on the correct class (i.e. $p=[0,...1,...,0]$ contains a single 1 at the $y_i$th position.). 

The Softmax loss is not only used in DCNNs, but also for training Feed-forward neural networks. In the task of image classification, many models use a softmax classifier for evaluation of loss, where as in the task of segmentation, multiple softmax classifiers are used which are centered at each spatial location on the output of the DCNN.

\subsubsection{Treating Convolution and pooling as Infinitely strong prior}
Prior acts as our assumed probability distribution over the parameter space for our model, based on our belief of what is reasonable, sans information about the data-sample at hand.A prior can be considered infinitely strong if it strictly forbids change of some parameter values, regardless of the support that the data provides. 

Now, a convolutional layer can be considered to be a Fully connected layer with some infinitely strong prior on its weights. The prior strictly forbids the network from having non-zero weights at locations other than a small contiguous area centered spatially at the unit called the receptive field of the unit.It also provides a strong prior in the form of ensuring that the weights for each unit in a layer must  be identical to its neighbors. Due to these priors, we are able to ensure that the function at each layer learns only local interactions and is equivariant to translation. Similarly pooling is equivalent to a strong prior that each unit should be invariant to small input translations~\cite{deeplearningbook}. 

As these are very strong priors, models using convolution and pooling may cause under fitting where such prior assumptions are not reasonably accurate. As we will see in the next section, these priors lead to models which slightly under perform in the task of semantic segmentation. For this task, additional modifications are included in DCNNs to include global context which is otherwise not valued due to these priors.

\subsubsection{Popular DCNN Architectures}
We will review some the most popular DCNN architectures which have been seminal in nature. This will provide an overview of the growth of DCNNs.
\begin{itemize}
\item \textbf{Lenet~\cite{lenet}} : LeNet is the first successful applications of Convolutional Networks.One LeNet architecture was used to read zip codes, digits, etc. This architecture is one of the best known LeNet architectures.
\item \textbf{AlexNet~\cite{alexnet}}. One of the most seminal work of Alex Krizhevsky, Ilya Sutskever and Geoff Hinton, the AlexNet, competed in the ImageNet ILSVRC challenge in 2012 significantly outperforming the second runner-up. A deeper and bigger version of LeNet in essence, AlexNet featured Convolutional Layers stacked on top of each other.
\item \textbf{ZF Net~\cite{zfnet}}. A Convolutional Network from Matthew Zeiler and Rob Fergus, which became known as the ZFNet, was the winner of ILSVRC 2013.They performed better than their competitors by improving AlexNet by tweaking the architecture hyperparameters, like the size of the middle convolutional layers and the stride and filter size on the first layer.
\item \textbf{GoogLeNet~\cite{googlelenet}} A Convolutional Network from Szegedy et al, from Google, was the winner of ILSVRC 2014. One key component they developed was the Inception Module that dramatically reduced the number of parameters in the network . Another major modification for the norm was the use of Average Pooling instead of Fully Connected layers at the top. This eliminated a large amount of parameters. Many follow-up versions to the GoogLeNet have been released, most recently Inception-v4~\cite{inceptionv4}.

\item \textbf{VGGNet~\cite{cnn_example2}} .Karen Simonyan and Andrew Zisserman developed a model now known as the VGGNet, which was the runner-up in ILSVRC 2014.The authors showed that for good performance, depth of the network was a critical component. Their final best network contains 16 CONV/FC layers and features an homogeneous architecture that only performs 3x3 convolutions and 2x2 pooling from the beginning to the end. Although a very popular network, two major problems with this network is that it uses a lot more memory and parameters and is additionally very expensive to compute. 

\item \textbf{ResNet~\cite{cnn_example3}}. Residual Network developed by Kaiming He et al., was the winner of ILSVRC 2015. It uses a lot of batch normalization~\cite{batchnorm} and a special skip connections between layers. The architecture did not include any fully connected layers at the end of the network.ResNets are currently the state-of-the-art models and are one of the primary choices for using ConvNets in real life applications.
\end{itemize}

\begin{table*}[thb]
	\label{tab:classification}
    \caption[Classification performance]{Comparison of performance of various Image classification Model on ImageNet Large Scale Visual Recognition Challenge (ILSVRC).}
    \centering{
	\begin{tabular}{| c | c | }
\hline
	\textbf{Model}& \textbf{Top-5 Test Error Rate } \\ \hline
	\textbf{AlexNet} & 15.3 \%  \\ \hline
	\textbf{ZF Net} & 14.8 \% \\ \hline
	\textbf{VGG Net} & 7.32 \% \\ \hline
	\textbf{Google LeNet} & 6.67 \% \\ \hline
	\textbf{ResNet} & 3.57 \% \\ \hline
	\end{tabular}
    }
\end{table*}

\subsection{Evaluation of Semantic Segmentation}
\subsubsection{Evaluation Metric}
Semantic segmentation is generally treated as a machine learning problem, where our model learns to predict the label at each pixel by being trained on training samples. Each of these training samples is provided along with a `ground truth image', which is the true segmentation of the sample, which we want the model to learn. The model is trained to minimize the difference between the prediction it gives and the ground truth. Once the model is trained we would like to evaluate its performance. 
However due to the common plague of over-fitting in machine learning, which leads to models that cannot perform well on images not used in training, we do not evaluate its performance not on the training set.

The performance is evaluated on a different fixed set of samples, known as test set, which has not been used in the training process. The performance on the test set shows the generalizability of the model. Aligning the training samples and test samples to have the same underlying distribution is crucial for gaining statistically significant results. 

For comparing the performance of various models and algorithms designed to tackle this problem, we can use many statistics such as mean per-pixel accuracy, clubbed per-pixel accuracy, Mean IOU.Primarily, the standard metric used to evaluate the algorithms is Mean IOU, which is the mean intersection over union.Mean IOU is evaluated by measuring the intersection and union between predicted region and actual ground truth region marked with the label for each label detected in the image.Mean IOU gives equal weightage to background objects, such as `roads', `grass' and foreground objects such as `cow’, `person’. A simple algorithm for calculating the mean IOU over a test set is given in Algorithm 2

\begin{algorithm}
\caption{Calculating the mean IOU on a testing set}\label{meaniou}
\begin{algorithmic}[2]
\State set mean IOU $= 0$ 
\For{Each label $l_i$ in dataset}
	\State $l_i$-Intersection $=$ 0
	\State $l_i$-Union $=$ 0

    \For{Each image I in test set}
        \State \textbf{Require :} True segmentation = $GT_I$.
        \State \textbf{Require :} Predicted segmentation = $P_I$.
        \State Intersection $=$ count(pixels marked $l_i$ in $GT_I$ \textbf{and} $P_I$)
        \State Union $=$ count(pixels marked $l_i$ in $GT_I$ \textbf{or} $P_I$)
        \State $l_i$-Intersection $+=$ Intersection
        \State $l_i$-Union $+=$ Union
	\EndFor
    \State  $l_i$-IOU $=$ $l_i$-Intersection $/$ $l_i$-Union
\EndFor
\State Mean IOU $ = \sum_i l_i$IOU / count(Labels)
\end{algorithmic}
\end{algorithm}

\subsubsection{Datasets}

Before the advent of deep learning models, datasets for evaluating semantic segmentation models used be be comparably smaller with less diversity. `Standford background dataset' was introduced in~\cite{stanforddataset}. It contains 715 images of 320-by240 pixel resolution, with 8 classes.`SIFT Flow dataset', introduced in ~\cite{siftflowdataset} contained 2,688 images and 33 labels. It is split into 2488 images for training and 200 for testing.An even wider dataset is the `Barcelona dataset', introduced in~\cite{barcelonadataset} contains 14,871 training images and 279 testing images. It contain 170 unique labels.Other datasets such as Berkeley segmentation dataset~\cite{berkleydataset}, known as BSDS500/BSDS300, and MSRC-21~\cite{msrc21} have also been used for comparing models.

One of the most popular datasets for evaluating semantic segmentation models is the PASCAL VOC 2012 dataset~\cite{pascalvoc2012}.The PASCAL VOC 2012 dataset includes 20 foreground classes and one background class. The original dataset contains 1464 training images, 1,449 validation images and 1456 test images.This dataset is augmented by~\cite{pascalaug} resulting in 10,582 training images. Other datasets such as `Coco'~\cite{coco}, `Cityscapes'~\cite{cityscapes} and `ADE20K'~\cite{ade20k} are also used for evaluation of model performance.

\section{Review of DCNN models for Semantic Segmentation}
This section contains review of DCNN models made for the task of semantic segmentation.Encapsulating every significant work in this field is not possible. Hence, this review has been made of cherry-picked DCNN models based on their performance as well as novelty. 

One of the earliest usage of DCNNs in the task of segmentation that became popular was in ~\cite{Hierar2013}, published in 2013. The proposed model uses multi-scale DCNN for extracting feature for each pixel and at varied scales to capture local and global information.In parallel, a Segmentation tree or superpixel based segmentation is computed. Now these two are combined in various fashion using techniques such as Conditional random Field. This model was able to out-perform contemporary models in `Stanford Background', `SIFT Flow' and `Barcelona' datasets.

Hiriharan et. al. introduced the concept of pixel hyper-columns for segmentation and fine-grained localization in~\cite{hypercolumns}, published in 2015.Arguing that only the top layer output of the DCNN is not the optimal representation of each pixel for fine-grained applications such as segmentation, they introduce the concept that the information of interest is spread over all layers of DCNN.A pixel's `Hypercolumn' is defined as outputs of all units above that pixel at all layers of the DCNN, stacked into one vector.These Hypercolumns are sparsely computed, as adjacent pixels would have strongly correlated Hypercolumns.The model gave state-of-the-art performance in the tasks of Simultaneous detection and segmentation(SDS) and locating keypoints.

The next seminal work in line was Fully convolutional networks\cite{fcn} by Johnathan Long et. al. With this, the concept of fully convolutional networks for segmentation was introduced.A skip architecture to combine deep, coarse semantic information with shallow, fine appearance information was also used.By converting the Fully connected layers to convolutional layers, the model ensures that spatial information is not thrown away.As the network subsamples to keep filters small and computational requirements reasonable, the network uses upsampling by bilinear interpolation, but also popularized the concept of `Deconvolution' to find segmentation at the same resolution as the input image.The network gave state-of-the-art performance on PASCAL VOC(mean IOU 67.5\% ), SIFT Flow, NYUDv2, and PASCAL-context. 

Many segmentation models face the task by modeling this problem as maximum a posteriori inference in a conditional random field defined over image patches or pixels.Before~\cite{crf}, fully connected CRFs were computationally very expensive to be utilized and hence, models used only partially connected CRFs. In this extraordinary work,Philipp Krahenbuhl and Vladlen Koltun introduced an highly efficient inference algorithm for fully connected CRF models, which reduced the time taken for fully connected CRF based inference exponentially. The paper was also rewarded the `Best student paper' award in `Nips, 2011', for its impact fullness. 
 
 	\begin{figure}[tbh]
		\label{fig:DeepLab Segmentation Model}
		\centering
		\includegraphics[scale=0.85]{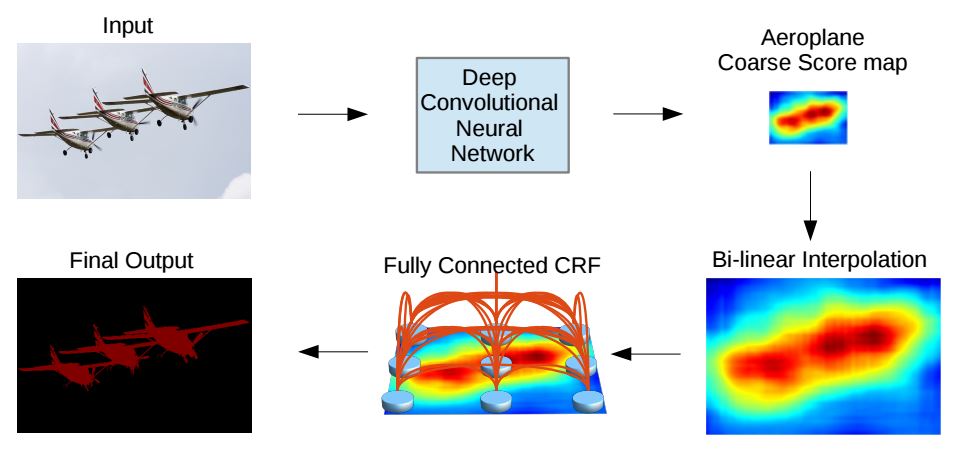}
        \caption[DeepLab Segmentation Network]{This figure shows the segmentation process of DeepLab Segmentation Networks.}
	\end{figure}

`DeepLab' was introduced in 2014 by Liang-Chien Chen et. al. in~\cite{deeplab2014}. `Deeplab' uses `atrous' convolution, which increases the receptive field of each unit in layer above without increasing the number of parameters. Due to the downsampling used in DCNNs, the output of a purely DCNN segmentation model lacks finer details. In `DeepLab', this is combated by uses fully connected CRFs with inference algorithm as in~\cite{crf}. The model uses an energy function : 
\begin{equation}
E(x) = \sum_{i}{\theta_i (x_i)} + \sum_{ij}{\theta_{ij}(x_i,x_j)}
\end{equation}
where $x$ is the label assignment for pixels.The unary potential is given by $\theta_i = -\log{P(x_i)}$, with $P(x_i)$ denoting the label assignment probability at pixel $i$.The pairwise potential is given by :
\begin{equation}
\theta_{ij}(x_i,x_j) = \mu(x_i,x_j) \sum_{m = 1}^{K} w_m k^m (f_i,f_j)
\end{equation}
where $\mu(x_i,x_j) = 1$ if $ x_i \neq x_j$ and zero otherwise.The gaussian kernels $k^m$ depends on features extracted for pixel $i$ and $j$. The kernels are :
\begin{equation}
w_1exp\bigg( -\frac{\norm{p_i-p_j}^2}{2\sigma_\alpha^2} - \frac{\norm{I_j-I_j}^2}{2\sigma_\beta^2} \bigg) + w_2 exp\bigg( - \frac{\norm{p_i - p_j}^2}{2\sigma_\gamma^2} \bigg) 
\end{equation}
where the hyperparameters $ \sigma_\gamma, \sigma_\beta, \sigma_\alpha$ control the scale of the gaussian kernels.The model was able to give state-ot-the-art performance in PASCAL VOC 2012 dataset(mean IOU 71.5\% ). 

In `DeepLab v2'~\cite{deeplabv2}, The model was further improved by using pre-trained `Residual Networks' for initialization. Additionally, inspired by~\cite{sppkaming}, the `atrous spatial pyramid pooling' was proposed, which extracted features at multiple parallel `atrous' convolution layers with different sampling rates, and combined them. The best variant of their model beat other state-of-the-art models in Cityscapes dataset, PASCAL-context dataset, and PASCAL VOC 2012 (mean IOU of 79.7 \%).

In Chronological order, `Mixed Context networks', introduced in ~\cite{hikseg},was the first model beat the performance of `DeepLab' on PASCAL VOC 2012 Dataset and MIT SceneParsing 150 dataset.The model mixes features at different `atrous' convolution rate, and learns to identify the most relevant scales. By combining `Deconvolution', as used in ~\cite{fcn} along with dilation with densely connected layers, which are finally processed by a `message passing module', the model was able to reach mean IOU of 81.4\% on PASCAL VOC 2012 dataset.
 
  \begin{figure}[tbh]
      \label{fig:Pyramidal Parsing Network}
      \centering
      \includegraphics[scale=0.75]{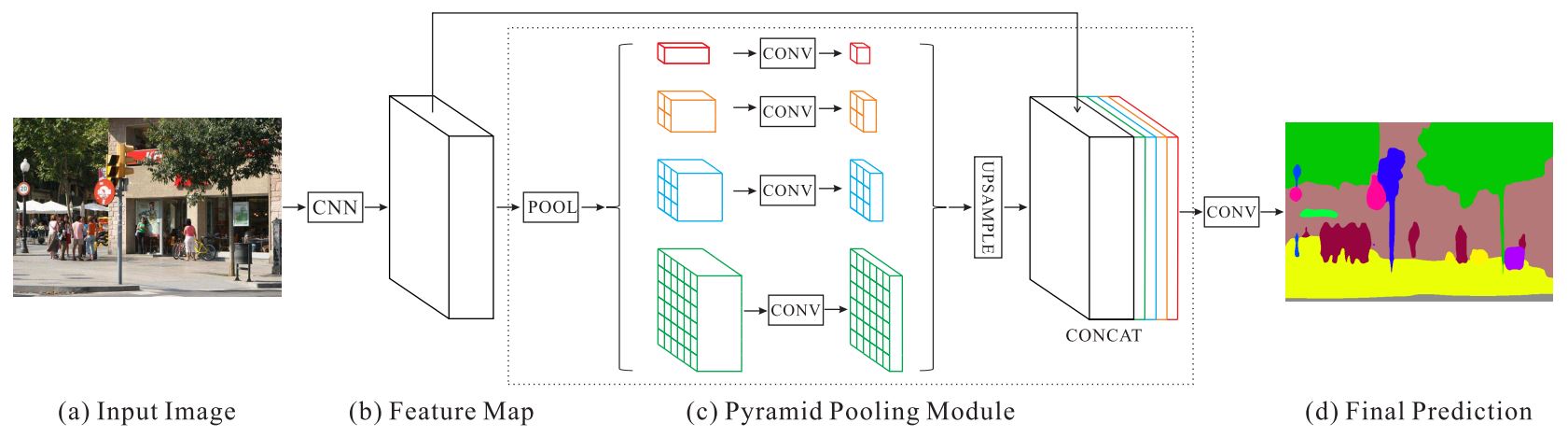}
      \caption[Pyramidal Parsing Network]{This figure shows the architecture of Pyramidal Parsing Network.}
  \end{figure}
  
Current state-of-the-art model is the `Pyramid Scene Parsing Network'~\cite{pspnet2016}. The model introduced pyramidal pooling module which exploits global context information by using different-region based context aggregation.Effecting the work introduced an effective optimization strategy for training Deep segmentation models.The pyramid pooling module is applied to the feature from the last layer of a DCNN, using which we extract sub-region representations, which are upsampled and concatenated for forming the input to another DCNN which gets the final per-pixel prediction.The State-of-the-art performance defined by this network stood at mean IOU of 85.4 \% on PASCAL VOC 2012 dataset, and 80.2 \% on Cityscapes dataset.

\begin{table*}[thb]
	\label{tab:Segmentation}
    \caption[Segmentation performance]{Comparison of performance of various Image Segmentation Models on PASCAL VOC 2012 Image Segmentation Challenge}
    \centering{
	\begin{tabular}{| c | c | }
\hline
	\textbf{Model}& \textbf{Mean IOU on 'test' Image Set } \\ \hline
	\textbf{Fully Convolutional Networks} & 67.5 \%  \\ \hline
	\textbf{DeepLab} & 71.5 \% \\ \hline
	\textbf{DeepLab v2} & 79.7 \% \\ \hline
	\textbf{Mixed Context Network} & 81.4 \% \\ \hline
	\textbf{Pyramidal Scene Parsing Network} & 85.4 \% \\ \hline
	\end{tabular}
    }
\end{table*}

\chapter{Segmentation Unravel}

\label{ch:Segmentation Unravel}

\section{Introduction : CNN Fixations}
\textbf{CNN Fixations is a work by Konda Reddy  Mopuri, Utsav Garg and R. Venkatesh Babu from VAL, IISc which is  currently under peer review. Due to the strong dependence of 'Segmentation Fixations' on this work, we explicitly explain the process of backtracking.}

Although DCNNs have demonstrated outstanding performance for recognition tasks such as hand written character recognition and object classification, they offer limited transparency.One way to understand CNNs is to look at the important image regions that influence it’s prediction.Such regions might also offer visual explanations in terms of the responsible image regions that misguide the CNN, when the predictions are not accurate.
 
  \begin{figure}[tbh]
      \label{fig:CNN Fixations}
      \centering
      \includegraphics[scale=0.45]{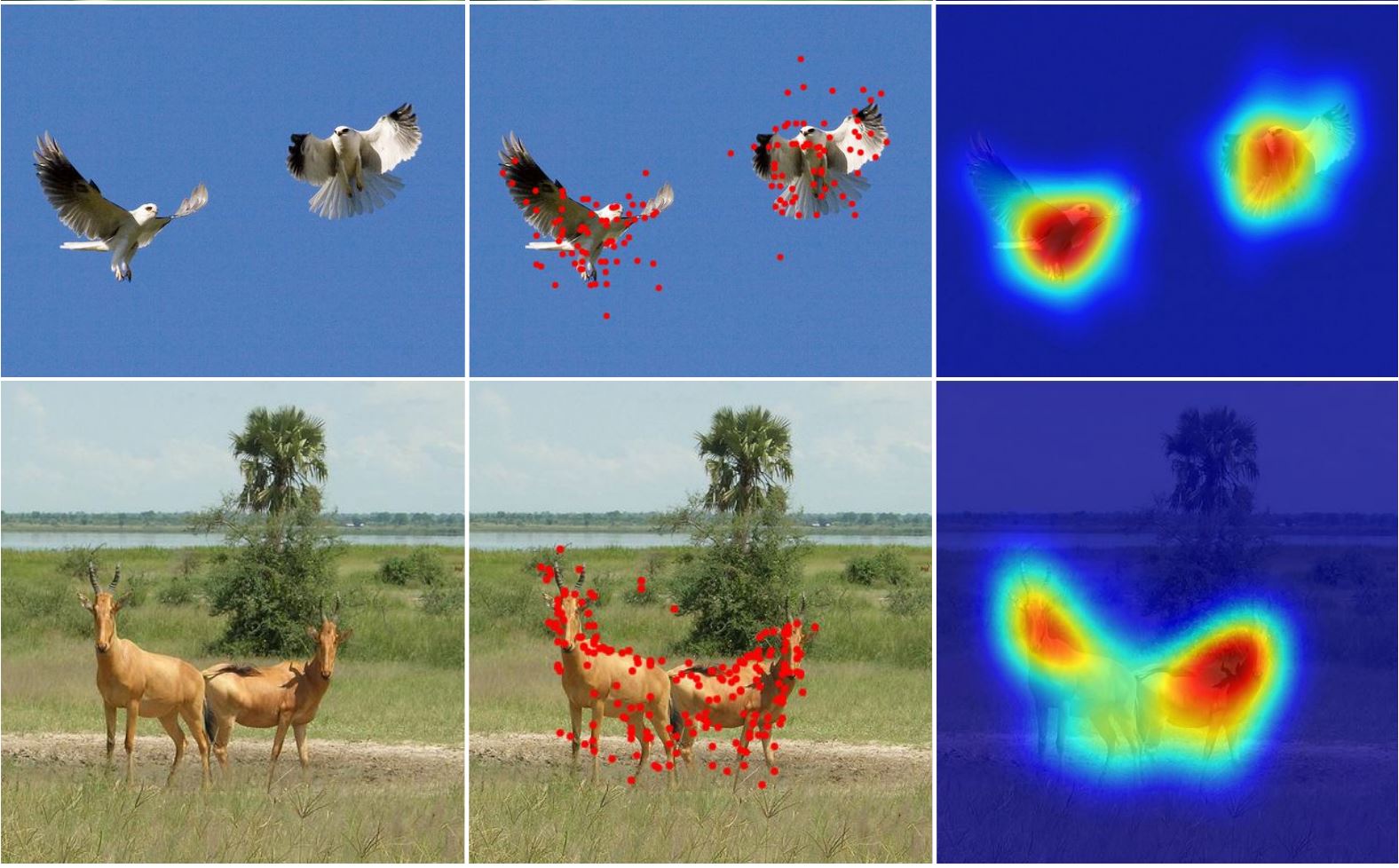}
      \caption[CNN Fixations]{Example CNN fixations. Each row shows an input image,
CNN fixations (red dots) determined by the proposed approach
and the density map obtained from the fixation points.}
  \end{figure}
In order to make the CNN based models more transparent and visualization friendly, they propose an approach to determine the important image regions that guide the model to its inference.As they are modeled analogous to human eye fixations, they are called CNN-fixations. The class specific discriminative regions are highlighted by tracing back the corresponding label activation via strong neuron activation paths to the image plane.

\subsection{CNN Fixation Approach}
Let $X_i$ be the feature map at $i$th layer,$i = 1, 2 ... M$ denotes the index of the layer in a CNN with a total of $M$ layers, $W_i$ denotes the weights connected to $i$th layer from it’s previous layer $i_1$, $B_i$ denotes the bias parameters at $i$th layer. $n_i$ denotes the number of neurons in $i$th layer, $f$ is non-linearity applied at each layer. They assume that the CNN is trained over $C$ object categories. In the proposed approach, they back-track the most active neuron from the deepest $(M)$ layer onto the image through all the intermediate layers.
  \begin{figure}[tbh]
      \label{fig:Proposed Approach}
      \centering
      \includegraphics[scale=0.75]{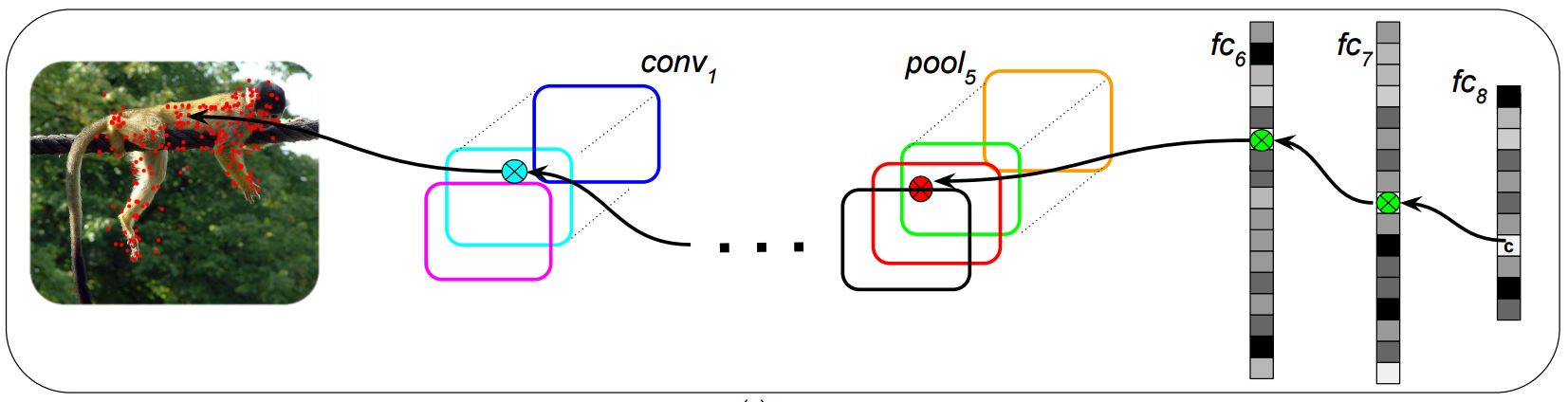}
      \caption[Unravelling in CNN Fixations]{Overview of the unraveling approach to obtain the CNN fixations}
  \end{figure}
  
The output at the $M$th layer, $X_M$, which denotes the predicted pre-softmax confidences towards the $C$ labels, is give by :
\begin{equation}
X_M = f(W_M . X_{M-1} + B_M)
\end{equation}

The predicted label $c$ is given by, $c = arg \max_j{X_M^j}$, where $X_M^j$ denotes the $j$th component of $X_M$. Now, $X_M^c$ can be written as 
\begin{equation}
X_M^c = f(<W_M^c, X_{M-1}> + B_M^c)
\end{equation}
where $W_M^c$ is the $c$th row of $X_M$ and $<.,.>$ denotes the inner product. In order to find the set of features in layer $M − 1$ that are
responsible for $X_M^c$ to fire, they observe the inner product term $< W_M^c, X_{M−1} >$. The indices of the top $K$ responsible features in layer $M − 1$, denoted as $I_{M−1}$, can be obtained using :
\begin{equation}
\label{fc-t-fc}
(I_{M-1}, V_{M-1}) = {top}_K(W_M^c \odot X_{M-1})
\end{equation}
where the function ${top}_K(z)$ outputs the indices $I$ of the top $K$ components $V$ in the vector $z$. They call this transition \textbf{fc-to-fc} transition. These act as seed of the layer $X_{M-1}$, from where they again find the ${top}_K$ contributors for these points by finding top contributors in the inner product $ (W_i^c \odot X_{i-1} )$. 
Once they have fixation points at Lowest fully connected layer, they need to find corresponding fixation points in the layer below it which might be a Convolutional layer or Pool layer.
For convolution,activations at each unit $(x,y,z)$, where $(x,y)$ corresponds to the spatial position and $(z)$ corresponds to the channel, is given by :
\begin{equation}
o(x,y,z) = f(W_M^{(z)} . X_{M-1}^{(x,y)} + B_M^{(z)})
\end{equation}
where, $W_M^{(z)}$ are the weights of the $(z)$th filter connection the layer below to layer above,$B_M^{(z)}$ is the corresponding bias for the $(z)$th filter, and $X_{M-1}^{(x,y)}$ corresponds the a tensor which consist of all the units in the receptive field of the unit $(x,y,z)$, belonging to the layer below. As the output of the activation is still a function of the inner product of a weight vector and an input vector, they follow the same procedure as in fc-to-fc transition. They call this \textbf{conv-to-conv} transition. The fixation  points in the layer below are given by \eqref{fc-t-fc}.

Across a Pool layer, the transition is much simpler. The output at location $(x,y,z)$ from a pool layer is given as : $o(x,y,z) = f(X_{M-1}^{(x,y,z)}) $, where $f$ is downsampling function like Average, or max, and $X_M^{(x,y,z)}$  corresponds the a tensor which consist of all the units in the receptive field of the unit $(x,y,z)$, belonging to the layer below in the channel $(z)$.Hence, Fixation point corresponding to each unit above is given by : 
\begin{equation}
(I_{M-1}, V_{M-1}) = {top}_K(X_{M-1}^{(x,y,z)})
\end{equation}
They apply these transition operations at each of the $M$ layers to reach to fixation points in the image corresponding to the predicted label.
\subsection{Conclusions}
The approach traces the evidence for a given neuron activation, in the preceding layers.CNN-fixations is a visualization technique, based on this approach.It highlights the image locations that are responsible for the predicted label. High resolution and discriminative localization maps are computed from these locations. In presence of multiple objects, they can sequentially discover individual objects and obtain their localization maps. 

\section{Extending CNN-fixations to Segmentation}

As seen in the previous section,CNN-fixations is a visualization technique for highlighting image locations responsible for the predicted label. In work for this thesis, this concept was extended to segmentation networks. The task now at hand is to highlight image regions which were responsible for the segmentation predicted by the network. It entails to finding, for each label detected,image regions which highly encouraged its labeling.

  \begin{figure}[tbh]
      \label{fig:segmentation Fixations}
      \centering
      \includegraphics[scale=0.50]{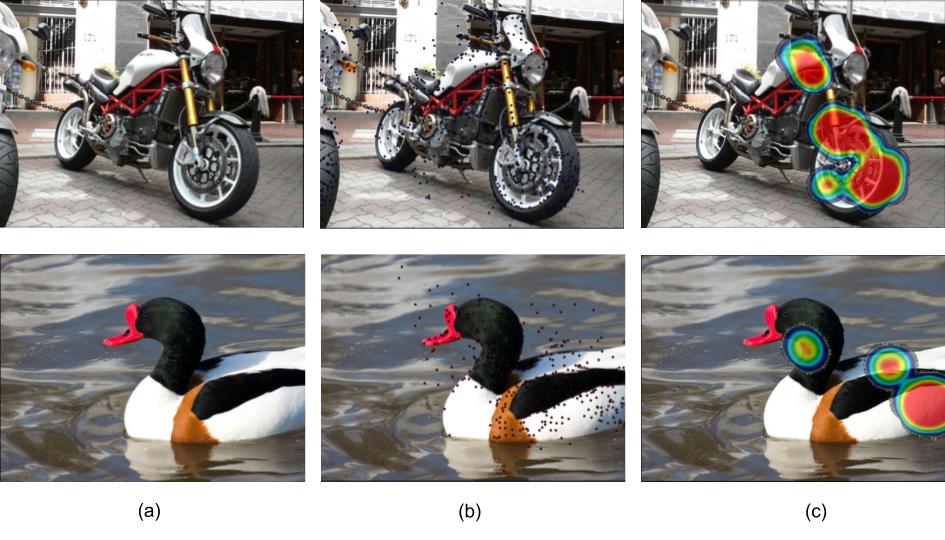}
      \caption[Segmentation Fixations]{Qualitative Examples of Segmentation Fixations. (a) The original Input Images,(b) The fixations found for the primary object in the image, (c) The heat map based on density of fixation points after removal of outliers}
  \end{figure}
  
Just as in CNN-fixations, we rely on the unraveling of strong neural activation pathways from the output to the input to find such regions. At each spatial location in the output, we have a predicted label. In models such as in~\cite{deeplab2014,pspnet2016}, the output is a tensor of shape $c\times h \times w$, where $c$ is the number of channels and $h$, $w$ corresponds to height and width respectively.The output blob contains, at each pixel, the activation value for each unique label for that pixel. This can be represented as a vector $Q(x,y) = \{ q_1(x,y),q_2(x,y), ...,q_c(x,y) \} $ of $c$ dimensions, which related to the probability distribution of each pixel by the equation :
\begin{equation}
P(l(x,y) = c_i) = \log{\frac{q_i(x,y)}{\sum_i{q_i(x,y)}}}
\end{equation}
where $l(x,y)$ is the label of pixel at position $(x,y)$.Each class is assigned the label which it has most probability of being. We take fixations for each label detected in the image at the output layer as the set of spatial points which were labeled as the object. 

  \begin{figure}[tbh]
      \label{fig:Segmentation Fixations In Details}
      \centering
      \includegraphics[scale=0.35]{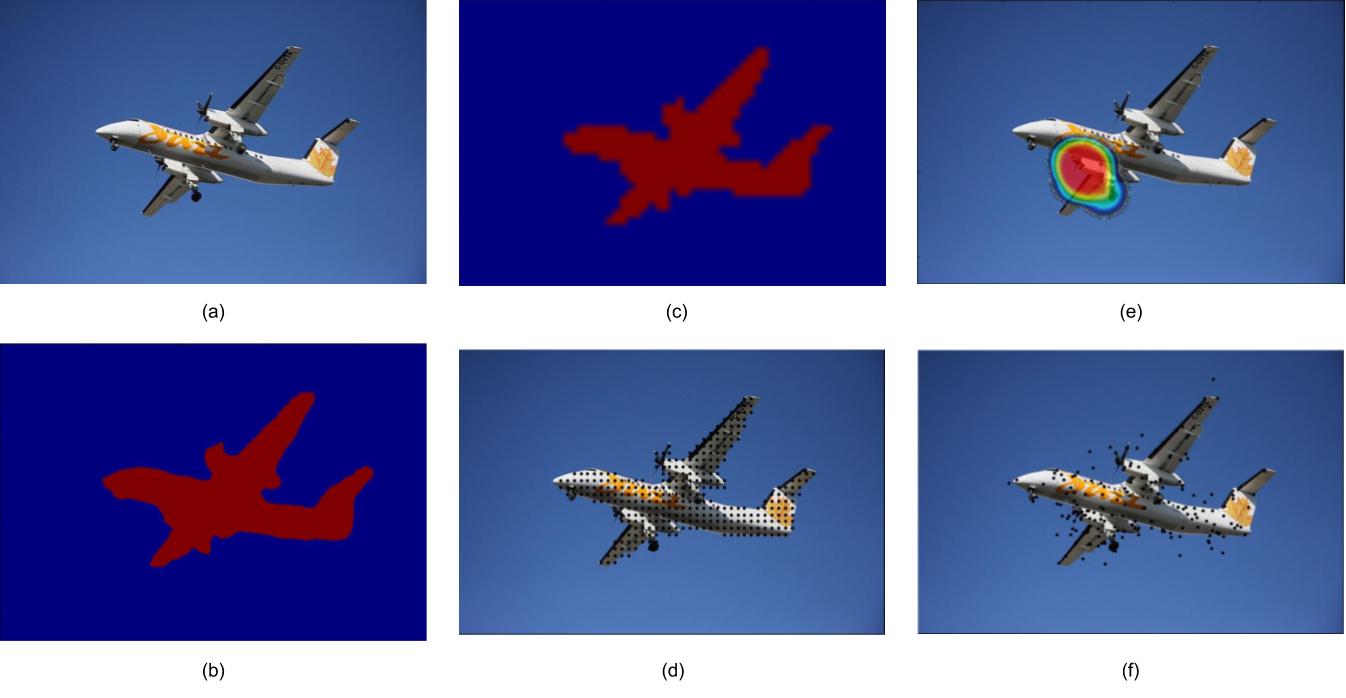}
      \caption[Segmentation Fixations In Details]{Segmentation Fixations In Details. (a) The original Input Images,(b) The Segmentation which is given as output from the model, (c) The downsampled output from argmax of activation layers, which is upsampled for (b), (d) Fixations \textbf{Initialized} at the top-most layer, (e)The Fixation heat map, (f) The fixations found for the primary object in the image.}
  \end{figure}

Each channel in the output corresponds to a unique labels. In the $c$ channel output tensor, these points are initialized at the corresponding class-layers. Once we receive the fixation points in the top layer, we use the similar transition algorithms between layers as in CNN-fixations to arrive at image-level fixations.These include conv-to-conv transition and pool transitions only as there are no fully connected layer in segmentation models. The prime differences between conv-to-conv transition in CNN-fixations and Segmentation-Fixations are :
\begin{itemize}
\item Allowing spatial shift of fixation points in conv-to-conv transition. When corresponding fixation point at two layers have do not have the same location in spatial dimensions(height and width), we say the fixation has went through a spatial shift.
\item an additional hyper-parameter `k', which is discussed in detail in the following subsection.
\end{itemize}
Some qualitative results of the fixations are given in figures.

  \begin{figure}[tbh]
      \label{fig:Segmentation Fixations Example}
      \centering
      \includegraphics[scale=0.275]{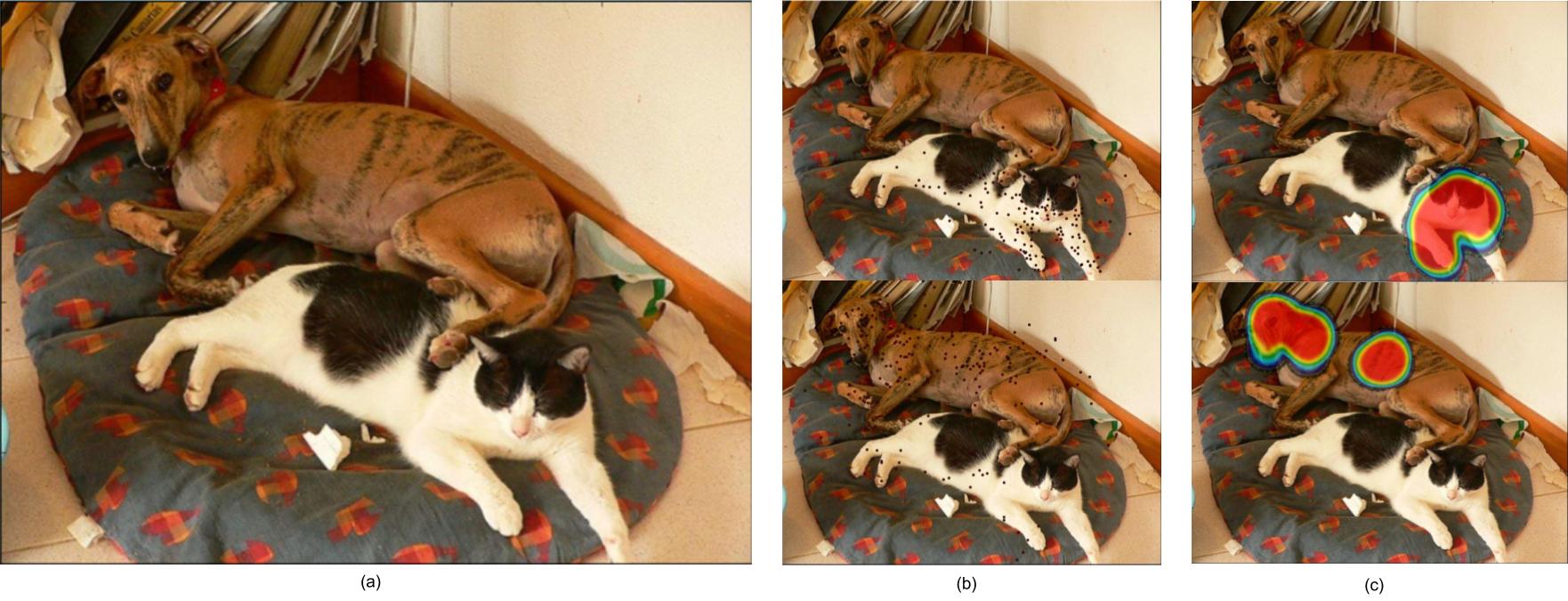}
      \caption[Multiple object fixations in Segmentation fixations]{Multiple objects in same image. (a) The original Input Images,(b) Fixations for `cat', and `dog' respectively, (c) The heatmap for salien region for `cat' and `dog'.}
  \end{figure}

Segmentation-Fixation finding models were made for Fully convolutional network~\cite{fcn}, DeepLab Large FOV(VGG-16)~\cite{deeplab2014} and Deeplab Multi-Scale `atrous' spatial pyramidal pooling Large FOV Residual Network model~\cite{deeplabv2}. These models were made on Caff~\cite{caffe}  and Pytorch. Some of the implementation use CPU parallel processing, as well as GPU Computation.

\subsection{K : An important Hyper-parameter}

While transitioning between layers, for each fixation point, we select the top `k' units which were connected to it as fixations on the layer below. Hence from each fixation point, while transitioning to the layer below, we have the potential to branch into multiple fixation points. The amount of branching out is regulated by the hyper-paramter `k'.

\subsection{Transition through `atrous' Convolution}
In networks with `atrous' convolution, the receptive field of each unit in top layers is enlarged significantly. To evaluate the effect of `atrous' convolution, we consider different variants of conv-to-conv transition in convolutional layers with 'atrous' convolution. We consider three types of conv-to-conv transition:
\begin{itemize}
\item \textbf{No Shift} : Spatial shift is not allowed in any convolutional layer transition
\item \textbf{Partial Shift} : We allow spatial shift at convolutional layers transition when the layer does not have 'atrous' convolution.
\item \textbf{Full Shift} : We allow spatial shift at every convolutional layer transition.
\end{itemize}

\section{Experiments}
For understanding Segementation-fixations we performed some experiments.These experiments were performed using DeepLab Large FOV model, based on VGG-16 with 16, containing 16 convolutional layers. The experiments were conducted on the 'val' set of PASCAL VOC 2012~\cite{pascalvoc2012}, which contains 1449 images.
\subsection{Experiment 1: Fixation Location}
For each image we find fixations for each object detected and evaluate how many lie inside the true segmentation of the object, using the ground truth. We calculate the average \% of fixation points of each object that lie inside the object across the dataset.

  \begin{figure}[tbh]
      \label{fig:Percentage Segmentation Fixations on object For different Object classes in PASCAL VOC2012}
      \centering
      \includegraphics[scale=0.65]{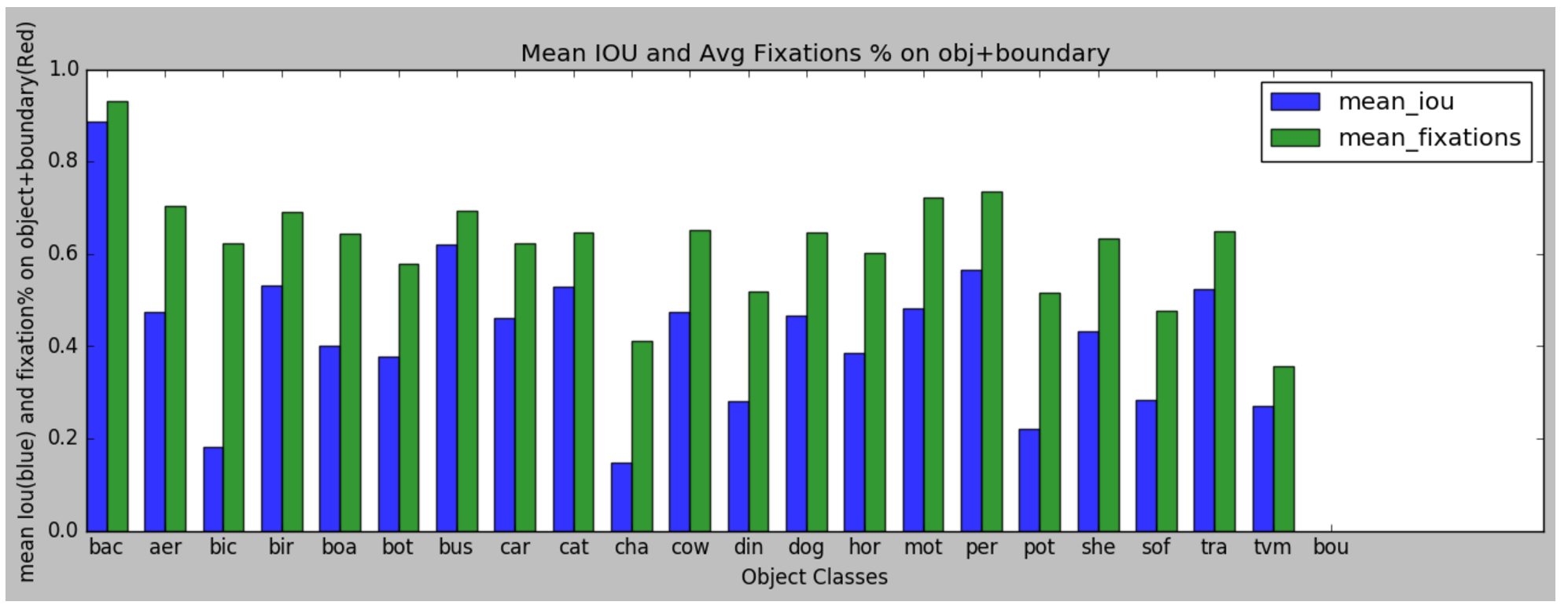}
      \caption[Segmentation Fixation quality evaulation]{ Percentage Segmentation Fixations on object For different Object classes in PASCAL VOC2012. Along with the percentage of fixations, we also show the average mean IOU for each class. }
  \end{figure}
  
The objects which have a poor segmentation performance also show a drop in the number of fixation points inside object. 

\subsection{Experiment 2: Effect of spatial shift variants}
To evaluate the effect atrous convolution on the strong neural activation pathways across layers, we evaluate the average \% of fixation points of each object that lie inside the object across the dataset using all the three different variants of spatial shift accross convolutional layers.

  \begin{figure}[tbh]
      \label{fig: Percentage Segmentation Fixations on object For different conv-transition strategies in PASCAL VOC2012}
      \centering
      \includegraphics[scale=0.65]{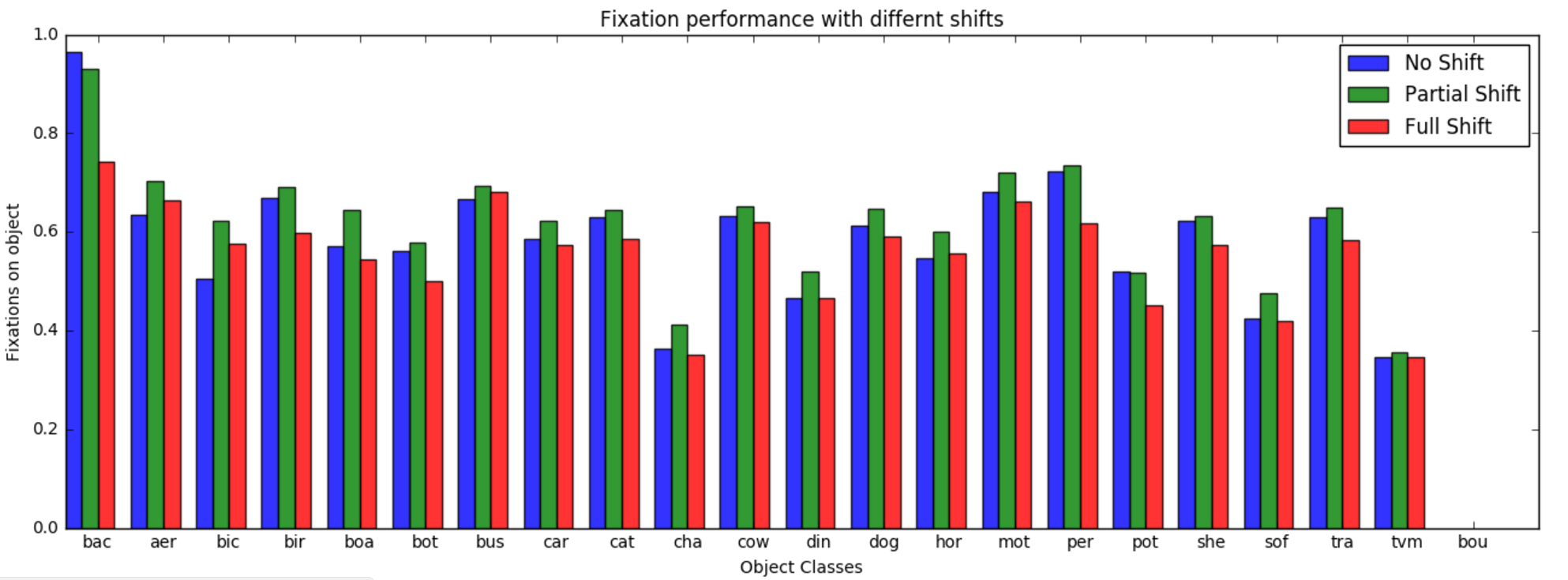}
      \caption[Effect of different conv-transition strategies]{Percentage Segmentation Fixations on object For different conv-transition strategies in PASCAL VOC2012 }
  \end{figure}

For the chart above we see that on an average by using `Full Shift' we land more fixation points outside the object than by using `Partial Shift' or `No Shift'. The variation in result shows that `atrous' convolution strongly influences the strong neural activation pathways across layers.The drop in \% also implies that the strongest activations were received from image regions beyond the object. 

In our objective of deriving inference of important image regions for object segmentation, we found fixation points found with `Partial Shift' appear more qualitative. When using `Partial Shift', fixation point seemed to latch to inner edges of the object rather than random edges outside the object.

\subsection{Experiment 3: Selecting `K'/ The distribution of activations in layers}
As discussed earlier, selecting the correct value for the hyper-parameter `K' at each layer is very crucial.
\begin{itemize}
\item Fixations are more indicative of important locations if they capture good amount of the activation for the prediction. This implies our process should not have too small `K'.
\item We will end up with a lot of fixations points which were very weakly influencing the outcome if our process has a bigger value of `K' than required. Hence we should not have too big values of `K'
\item Too big K will lead to heavy Computational Cost. The effect is exponential as the number of fixation points in image are $O(K^m)$ where $m$ is the number of layers.
\end{itemize}

Using more than 2,00,000 fixations points(5 images - across all layers) we find properties of activation distribution in layers to evaluate a strategy for finding `K' at each layer.At each fixation point at any layer of the network, we evaluate what the value of `K' must be to capture, $x$ \% of the total activation of the fixation point.This detail is averaged across all objects in all images. The result are shown in the plots below :

  \begin{figure}[tbh]
      \label{fig:`k' Required for capturing `x' Percent Activation}
      \centering
      \includegraphics[scale=0.5]{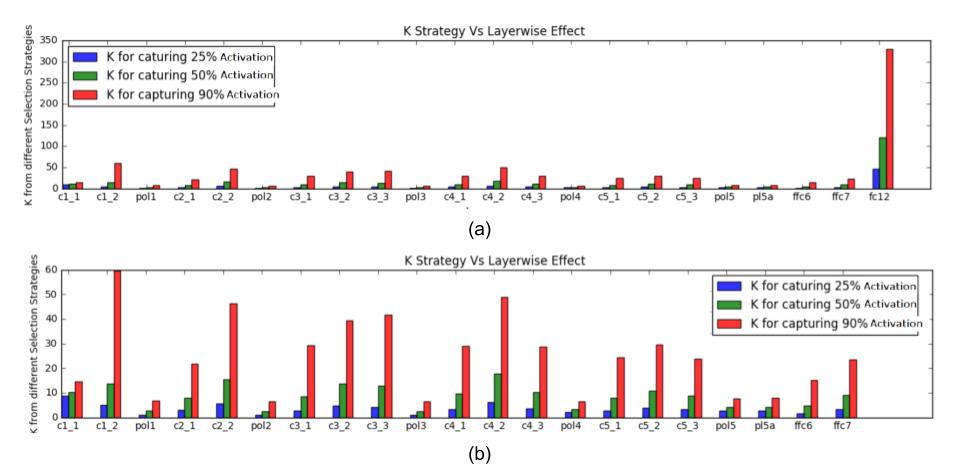}
      \caption[Analysis of required K across layers]{`K' Required for capturing `x' Percent Activation of unit above at various layers of DeepLab Large FOV (VGG-16) model.}
  \end{figure}

The spread of `K' showed that strategies such as `K' for capturing even 50 \% of the unit's activation were computationally too expensive. Such a strategy could lead to $15^m$ (m is number of layers) fixation points in the image plane from each fixation point initialized in the top layer.  This led us to use smaller strategies for `K', such as
\begin{itemize}
\item `K' $= 1$
\item 'K' such that it captures atleas 10\% of activation.
\item 'K' which increases according to the spatial resolution increase between layers.
\end{itemize}

Selecting the correct strategy for `K' leads us to more reliable segmentation Fixations. Results from using the third strategy captured sufficient information at decent computational cost.

\section{Salient parts for segmentation}

As discussed earlier, Convolution and pooling can be considered as a strong prior on Feed-forward networks. This prior might make the model under performs when the prior is not reasonable. In the task of semantic segmentation, Global context plays a crucial role in providing accurate segmentation predictions. As scene in Section 1.2.3, various models cope with this problem by adding various other modules and variations in convolutions to provide global context to the classifier at each pixel.

To better understand this need of global context in the task at hand, we perform the following experiment.

\subsection{Experiment 4: Existence of Salient Parts}
 On various images from the dataset as well as unseen data, we find the segmentation twice, once of the whole image, and once by masking discriminative regions of the object.The segmentation is found using DeepLab Large FOV(VGG-16) model.
\newpage
  \begin{figure}[tbh]
      \label{fig:Masked Object Segmentation}
      \centering
      \includegraphics[scale=0.37]{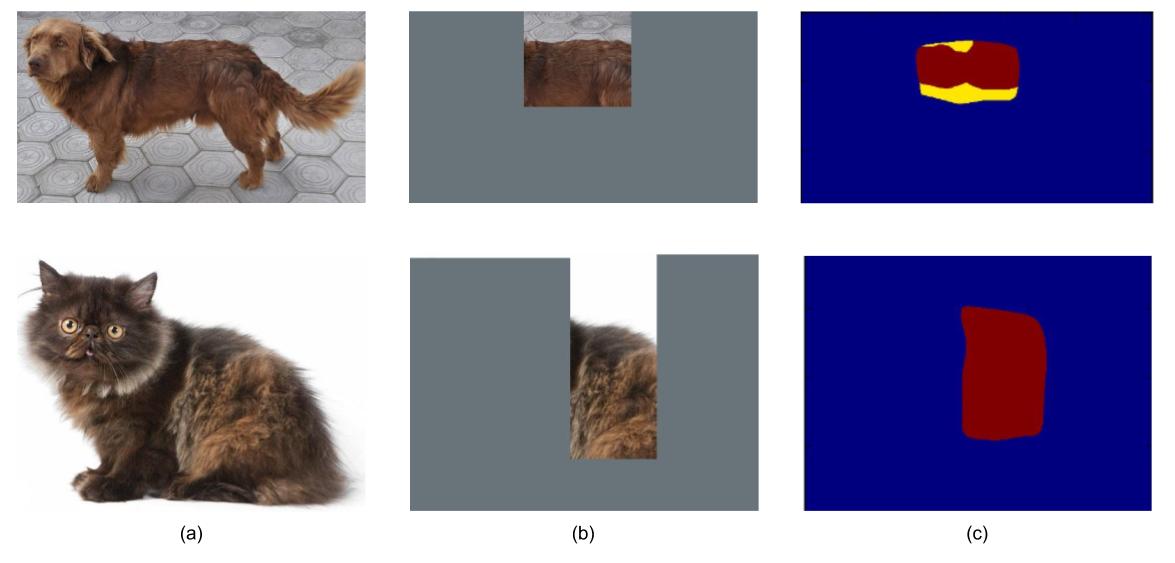}
      \caption[Masked object Segmentation]{Masked object Segmentation: (a) The original input images from unseen data(Google), (b) The masked images so that discriminative regions of object(For eg, Face) are covered, (c) Map showing output segmentation of masked Images. Blue corresponds to label 'Background', Brown in images at top and bottom corresponds to `Dog' and `Cat' repectively. }
  \end{figure}

From the above results we see that the model was able to capture the true segmentation in most of the cases even after hiding discriminative regions. Does it mean that global context is not important?

For finding the answer to this question we extended the experiment. The activation for the object on the masked image is compared with the activation for the object on the original image.

  \begin{figure}[tbh]
      \label{fig:Masked Object Segmentation: Importance of Discriminative regions.}
      \centering
      \includegraphics[scale=0.23]{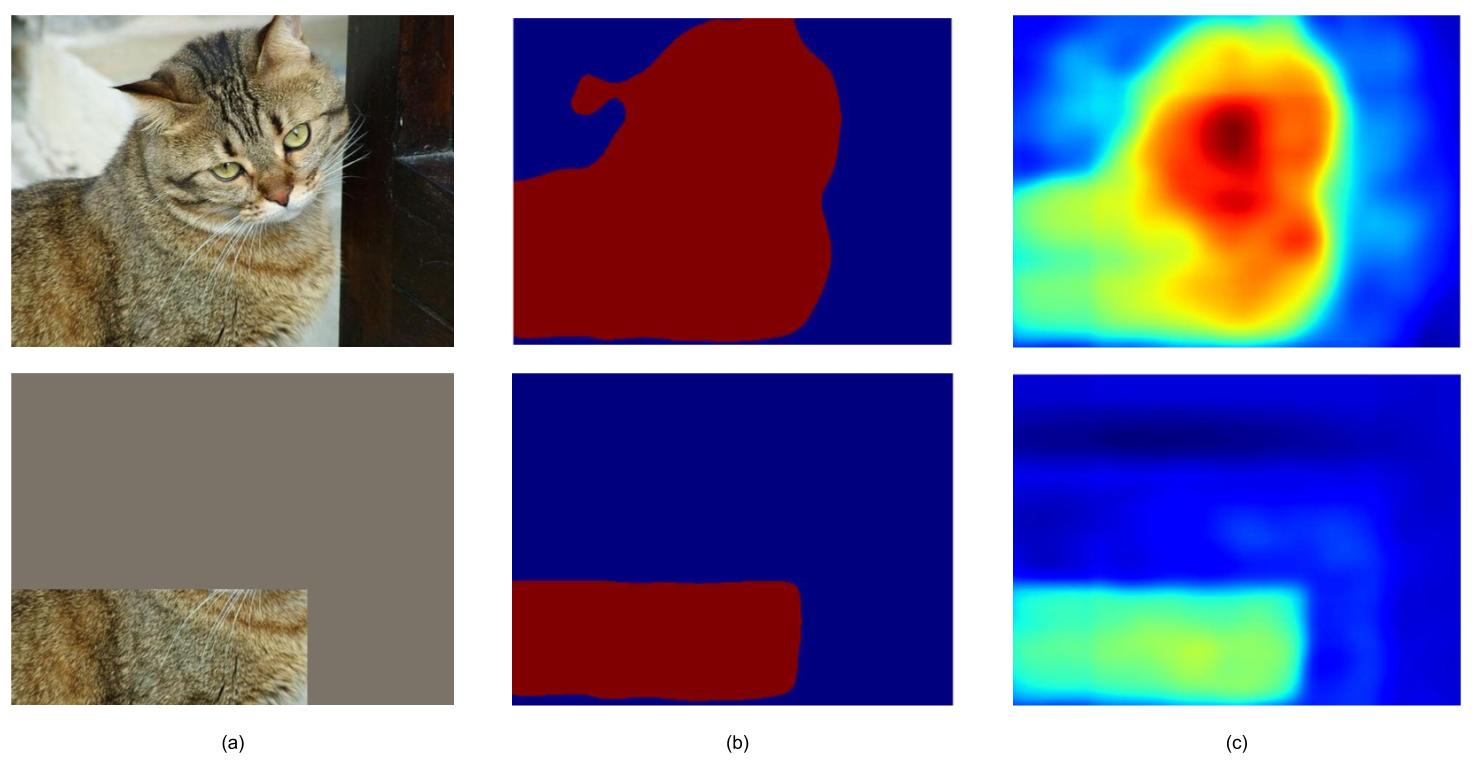}
      \caption[Masked Object Segmentation with activation map]{Masked Object Segmentation: Importance of Discriminative regions: (a) Original Image and Masked Image, (b) Predicted segmentation from network(Brown = `Cat'), (c) Activation map at channel corresponding to `Cat' object.}
  \end{figure}

In the above example it is seen that the presence of discriminative region enhances the activation for the object across the image. This result shows that high activation `flows' from the discriminative regions of the object to the other parts. Based on this experiment we create a class of Auxiliary loss functions which help in improving segmentation by encouraging the `flow' of activation across the object from the discriminative region. 

  \begin{figure}[tbh]
      \label{fig:Masked Object Segmentation: Activation increase by Discriminative regions.}
      \centering
      \includegraphics[scale=0.4]{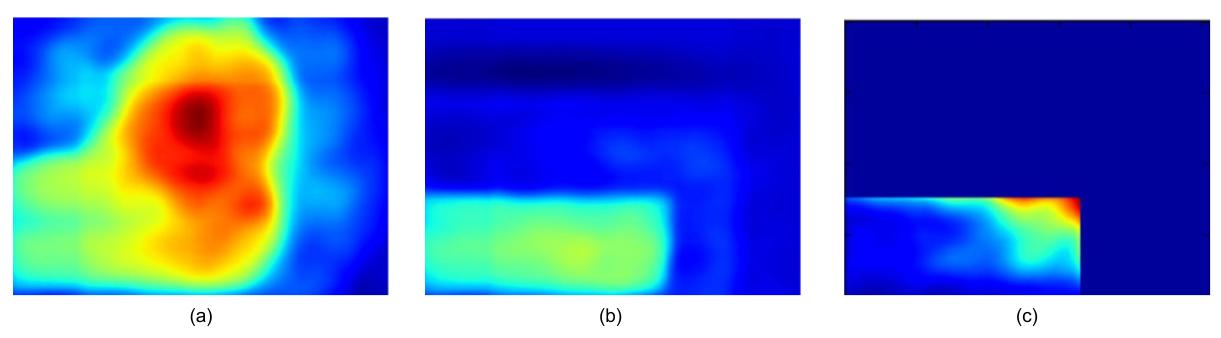}
      \caption[Activation map enhancement by discriminative features]{Masked Object Segmentation: Activation increase by Discriminative regions. (a) The activation map `Cat' layer in segmentation of the original Image , (b) The activation map `Cat' layer in segmentation of the masked Image, (c)  The additional activation gained at the unmasked location due to the presence of discriminative region.}
  \end{figure}
%




\chapter{Per-Pixel Feedback in Auxiliary Loss}
\label{ch:aux_loss}

\section{Introduction}
As discussed in Section 2.4, presence of discriminative regions of object, boost the activation for the object across the object segmentation in the image. This shows that a successful segmentation model will capture the presence of discriminative region, and propagate additional activation for the object to all regions of the object. We introduce a family of loss function with this goal in mind. 

Additionally, there are many unique labels which end up with highly correlated internal abstract representations in the model. Often objects of one of the correlated label is partially segmented as the object it is correlated to. This family of loss function additionally tries to decrease this similarity.

This family of loss functions is modeled to :
\begin{itemize}
\item Enhance segmentation using global context
\item Improve the networks capability to distinguish between highly correlated classes such as `cat' and `dog', `sofa' and `chair'.
\end{itemize}

\section{Per-pixel Template-similarity loss}
In many architecture, the output is a $c$ channeled output, where each channel contains a spatial map of the activations for a unique label. This output is generally generated by a convolution layer which gives $c$ outputs, and uses a $1 \times 1 \times l$ kernel, where $l$ represents number of channels in the output of the layer just before the final layer.

Lets consider a (VGG-16) based segmentation model trained to give a 21-class prediction as in PASCAL VOC 2012. The model has on the top a layer `fc8' which gives as an output a $O$, which is a $21 \times h \times w$ tensor, which corresponsds to downsampled activations for objects at various spatial positions. The layer before it, `fc7' gives as an output $O_l$, a $1024 \times h\times w$ tensor, which is taken as input by `fc8' layer.

As the model finally uses only a $1 \times 1$ kernel at each spatial position, it makes inference about the label of the unit by only relying on the $1024$ dimensional representation$o_l(x,y)$ of the spatial position in output $O_l$ of the `fc7' layer.

  \begin{figure}[tbh]
      \label{fig:Auxiliary Loss : Representation at second last layer.}
      \centering
      \includegraphics[scale=0.35]{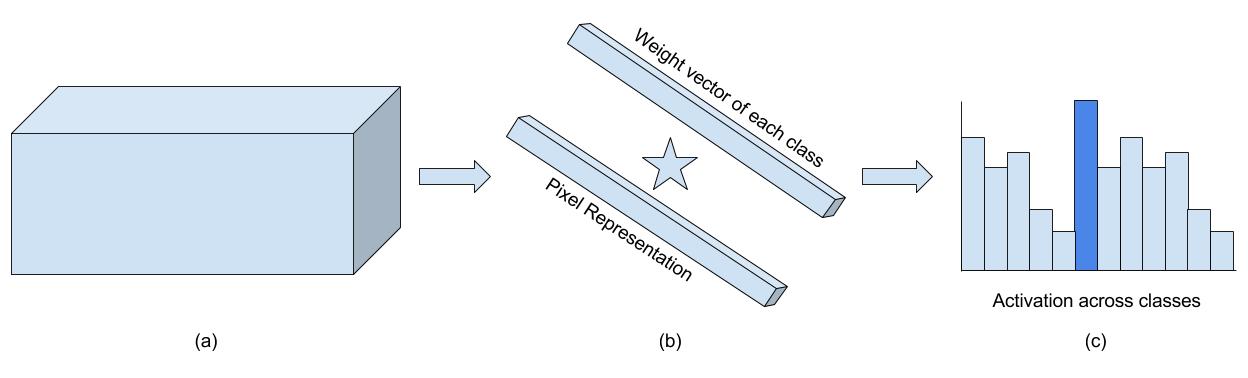}
      \caption[Auxiliary Loss : Representation at second last layer.]{Auxiliary Loss : Representation at second last layer. (a) The `fc7' layer has a $1024 \times h\times w$ tensor, which is connected to `fc8' by $1\times 1$ kernel convolution. (b) At each pixel, Activation for each class is found by inner product of corresponding weight vector and the $1024$ dimensional representation of the pixel. (c) The pixel is assinged the class label for which it has the highest activation.}
  \end{figure}

Now, this $1024$ dimensional representation of each spatial position, given by $o_l(x,y)$ contains all the information required for the network to classify it. All the additional information gained through various means such as pooling across scales or `atrous' convolution has now been encoded in this $o_l(x,y)$ representation of the spatial position.

Now, for each object detected in the image, we select the best $1024$ dimensional template $o_l{x^*,y^*}$ of it. After selecting the best template for each class we define a loss based on the similarity of each pixel to these representation. If the object belongs to the same class, the similarity is treated as a profit. If the object of a different class, the similarity is treated as a loss. A detailed algorithm is provided below :

  \begin{figure}[tbh]
      \label{fig:How our per-pixel Auxiliary Loss works}
      \centering
      \includegraphics[scale=0.35]{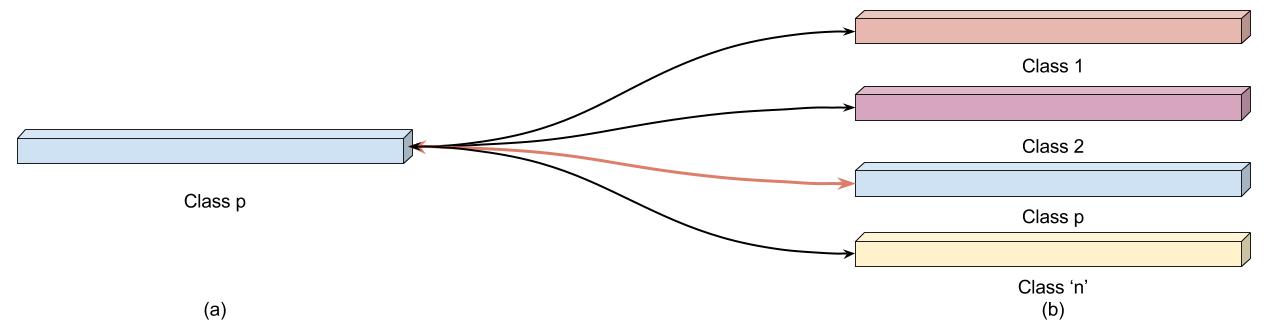}
      \caption[Per-Pixel Auxiliary Loss evaluation]{How our per-pixel Auxiliary Loss works: (a) For each pixel, we find its $1024$ dimensional representation. (b) The similarity of this pixel with the best template of each detected class is computed. If the template and pixel have different classes, similarity is a loss, else it is a profit.}
  \end{figure}

\begin{algorithm}
\caption{Per-pixel template similarity loss a batch of images while training}\label{corloss}
\begin{algorithmic}
\State \textbf{Require :} representaion tensor = $input$
\State \textbf{Require :} Ground Truth  = $GT$
\State Find all detected labels in batch.
\State loss = 0
\For{each label c in batch}
	\State $loss_c = 0$
    \State $mask = binary(GT = label)$
    \State ${Input}_{masked} = $ $input \times mask$
	\State $o_l(x^*,y^*) = $ best template for c in ${Input}_{masked}$
	\For{ each pixel in batch}
    	\State $o_l(x,y) = $ representation of pixel
    	\State $sim =$ Similarity($o_l(x,y) , o_l(x^*,y^*)$)
	    \If{$label(pixel) = c$}
	    	\State $sim = -sim$
	    \EndIf
	    \State $loss_c += sim$
    \EndFor
	\State $loss += loss_c$
\EndFor
\State $loss = loss / count($detected labels in batch$)$
\end{algorithmic}
\end{algorithm}

\section{Some Variations of Per-pixel Template-similarity loss}
Per-pixel Template-similarity loss can have many variations. There are two important factors :
\begin{itemize}
\item How we select the best representation of each template.
\item How we measure the similarity between different representations.
\end{itemize}
 Based on different strategies for these two factors, we can get a family of Per-pixel Template-similarity losses. We list two that we used for our experiments.
 \subsection{Correlation Loss}
 For each label detected in batch, we find the best representation $o_l(x^*,y^*)$ by finding the $O_l$ representation of the the spatial position with the highest activation of the label in $O$. As correlation is a good measure of how two set of vectors are related, we measure similarity by using the correlation between $o_l(x^*,y^*) = X*$ and $o_l(x,y) = X$ at each pixel. This is given by :
 \begin{equation}
 cor(X,X*) = \frac{<X*,X> - \mu_{X} \mu_{X^*} }{\sigma_{X^*} \quad \sigma_{X}}
 \end{equation}
We call a minor variant of this loss `selective correlation loss', where loss is computed only at the misclassified labels.
 
 \subsection{Cosine Loss}
Similar to the `Correlation Loss', for each label detected in batch, we find the best representation $o_l(x^*,y^*)$ by finding the $O_l$ representation of the the spatial position with the highest activation of the label in $O$. The Cosine Distance between two vectors measure the angle them. Hence we measure similarity by using the Cosine Distance between $o_l(x^*,y^*) = X*$ and $o_l(x,y) = X$ at each pixel. This is given by :
 \begin{equation}
 cos(X,X*) = \frac{<X*,X>}{\norm{X^*} \norm{X}}
 \end{equation}
 A minor variant of this loss is `selective cosine loss', where loss is computed only at the misclassified labels.
 
\section{Experiments}
We use a pre-trained `Deeplab Large FOV (VGG-16)' model. The model is trained for fine-tuning with correlation, cosine, selective correlation and selective cosine loss. The training is performed in augmentation of PASCAL VOC 2012 'train' dataset, which contains 1456 images. We set learning rate of Learning rate for all layers except the last three to 0. The learning rate is reduced to 0.0001, and the Learning rate reduced by polynomial learning policy.

We train the model along with Softmax loss. These losses are used as auxiliary loss functions to fine-tune the model. Various values of the hyper-parameter Correlation weighting factor  are used to evaluate which performs the best.


\begin{table*}[thb]
	\label{tab:Auxiliary loss performance}
    \centering{
	\begin{tabular}{| c | c | c | c | c | c | c |}
	\hline
	\textbf{Auxiliary loss}     &\multicolumn{6}{ c |}{Earlier Performance = \textbf{0.6225}}      \\ 
    \cline{2-7} 
    & \multicolumn{6}{ c |}{\textbf{Hyper Parameter Values}}
\\ 
	\cline{2-7}
	& \textbf{0.01} & \textbf{0.05} & \textbf{0.1} & \textbf{0.5} &\textbf{1}&\textbf{5}\\
\hline

	Correlation Loss & \textbf{0.6314} & 0.6285 & 0.6252 & 00.5977& 0.5560& 0.3253 \\ \hline
	Cosine Loss& 0.6331 & 0.6306 &\textbf{0.6321}& 0.6198 & 0.6086 & 0.5996\\ \hline
	Selective Correlation Loss & \textbf{0.62645} & 0.32536 & 0.32535 &0.32535 & 0.32535& 0.32535\\ \hline
	Selective Cosine Loss& 0.63324 & \textbf{0.63767} & 0.63104 & 0.61689 & 0.60718& 0.59921\\ \hline
	\end{tabular}
    }
    \caption[Auxiliary Loss performance]{Auxiliary Loss performance : Comparison of performance of Model with various weight factor hyper-parameter on various Per-pixel template similarity loss}
\end{table*}

\subsection{Conclusion}
As scene from table 3.1, Per-pixel template similarity loss are able to improve performance of Deeplab Large FOV(VGG-16) model at small values of hyper-parameters. However, an extensive statistical evidence for the usefulness of such loss functions must be collected across a plethora of segmentation models.

The follow-up work will try to find stronger statistical evidence for Per-pixel template similarity losses. Major issues such as quality of enhancement on models with high performance, effect of training routines and better similarity measurement will be explored.


\chapter{Conclusion}

In this thesis,we have extended the CNN-Fixations work which provides visual explanations for CNN predictions to semantic segmentation. We find such regions through `Segmentation-Fixations', which essentially backtrack strong neural activation pathways to salient image regions.We conduct various experiments to understand the best settings for `Semantic Segmentation'. Further more, we explore the importance of such salient regions in this task, and introduce the concept of `flow of activation', from discriminative object regions to the all regions of the object.

Based on the various experiments, we propose a family of loss functions to fine-tune the performance of Semantic-segmentation models. We call these `Per-pixel template similarity' losses. We show improvement in performance of `DeepLab Large FOV(VGG-16)' model on PASCAL VOC 2012 dataset by using variants of `Per-pixel template similarity'. However, further test remain to conclusively evaluate the benefits due to the proposed loss functions across models, and across data-sets.

Future work will be focused on statistically validating the benefits of the proposed loss functions.

\appendix

\addcontentsline{toc}{chapter}{Bibliography}
\bibliographystyle{plain}
\bibliography{bibliography/introduction}

\end{document}